\documentclass[preprint,12pt]{elsarticle}
\usepackage[a4paper,left=1in,right=1in,top=1in,bottom=1in]{geometry}
\usepackage{amssymb}
\usepackage{algorithmicx}
\usepackage{algpseudocode}
\usepackage{algorithm, float}
\usepackage{graphicx}
\usepackage{adjustbox}
\usepackage{setspace}
\usepackage{booktabs}
\usepackage{multirow}
\usepackage{siunitx}
\usepackage[cmex10]{amsmath}
\usepackage{multirow}
\journal{*}

\begin{document}

\begin{frontmatter}



\title{An Automatic Contextual Analysis and Clustering Classifiers Ensemble approach to Sentiment Analysis}


\author{Murtadha Talib AL-Sharuee\corref{cor1}}
\cortext[cor1]{Corresponding author. }
\ead{al-sharuee.m@students.latrobe.edu.au}
\author{Fei Liu\corref{}}
\author{Mahardhika Pratama\corref{}}
\address{Department of Computer Science and Information Technology, La Trobe University, Bundoora, Victoria 3086, AUSTRALIA}

\begin{abstract}
%^
Products reviews are one of the major resources to determine the public sentiment. The existing literature on reviews sentiment analysis mainly utilizes supervised paradigm, which needs labeled data to be trained on and suffers from domain-dependency. This article addresses these issues by describes a completely automatic approach for sentiment analysis based on unsupervised ensemble learning. The method consists of two phases. The first phase is contextual analysis, which has five processes, namely (1) data preparation; (2) spelling correction; (3) intensifier handling; (4) negation handling and (5) contrast handling. The second phase comprises the unsupervised learning approach, which is an ensemble of clustering classifiers using a majority voting mechanism with different weight schemes. The base classifier of the ensemble method is a modified k-means algorithm. The base classifier is modified by extracting initial centroids from the feature set via using SentWordNet (SWN). We also introduce new sentiment analysis problems of Australian airlines and home builders which offer potential benchmark problems in the sentiment analysis field. Our experiments on datasets from different domains show that contextual analysis and the ensemble phases improve the clustering performance in term of accuracy, stability and generalization ability.

\end{abstract}


\end{frontmatter}


\section{Introduction}
\label{sec1}

Sentiment analysis (SA) is a computational analysis of user-generated materials, such as reviews, to determine its orientation (positive, negative or neutral). There are two main reasons to automate sentiment classification: first, the abundance of online materials, which is the result of web development, is beyond human analysis; and second, public opinion is a significant consideration when governments, institutions and individuals are making decisions and taking actions. Many diverse domains and applications can benefit from SA, including those in the political \citep{tumasjan2010predicting,kato2008taking} linguistic \citep{guerini2016urban} medical and social issues \citep{pestian2012sentiment} and financial \citep{peetz2015estimating,smailovic2014stream,sehgal2007sops} domains. Thus a considerable attention has been drawn to SA and closely-related research directions such as emotion detection \citep{li2014text,balahur2013detecting}, subjectivity analysis \citep{mangassarian2007general}, irony detection \citep{reyes2012humor,reyes2011mining}, and contention texts analysis \citep{trabelsi2015extraction}. SA has been the focus of computer science research for more than 15 years and has been a topic of over 7,000 research works \citep{feldman2013techniques}. A large number of machine learning approaches have been proposed for classifying text in terms of the sentiment it expresses.
However, most of proposed methodologies were based on supervised paradigm which requires either pre-training or human participation during the classification. Whereas unsupervised methods, which are completed automatic, do not normally achieve an accuracy at a satisfactory level. These seriously affected their usability and effectiveness, especially in handling online data which may require constant training and human participation.

This article introduces an unsupervised and completely automatic method for classifing reviews, which consists of two phases, contextual analysis and clustering classifier ensemble. The first phase enables an automatic contextual analysis by effectively deploying a sentiment lexicon, SentiWordNet 3.0 (SWN), to prepare text for further processing and to address common linguistic forms which are intensifiers, negation and contrast. Addrissing sentiment modifiers is substantial procedure in our algorithm because they are very common forms and they lead to a significant sentiment modification. For example, the sentence "It is not a good movie" will be considered as a positive expression if the negation is not taken into consideration. Contrast also affects the overall polarity, for instance, the sentence "very long movie, however, we enjoyed it", in our method it is assumed that the second part of the sentence "we enjoyed it" is the overall author's sentiment, which should be emphasized on when classifing text. The second phase of the proposed method is a binary classification by applying an ensemble learning method with modified k-means classifier as a component learner. The ensemble learning is meant to improve the clustering result because it handles the bias-and-variance problem better than a single model approach. After dealing with polarity shifters, the adjectives and adverbs in all the documents are extracted as a set of features. Thereafter we operate k-means on several vector space models with different weights schemes which will be combined using voting mechanism. To develop a reliable method, SWN , is utilized to cluster the feature set into three groups positive/negative and neutral. Then the positive and negative features groups are set to be two initial centroids (initial seeds) of k-means. A positive document which is composed of all positive features and a negative document which is composed of all negative features will also be used, later on, for group judgment. 
%^

The main contribution of the research is that it defines a completely automatic unsupervised method for sentiment classification, which requires no training or human participation. As a result, the proposed method is a domain-independent approach to SA and particularly effective in processing a high volume data. The unsupervised approach also aims to produce a labelling-free model which is highly desired in practice due to its low manual intervention. As a contextual analysis and clustering algorithm, it is suitable for sentiment classification because usually, in the real world, the actual need is to analyse a large quantity of reviews, not to predict a single or few instances.  However, few studies have introduced unsupervised clustering algorithms to the field of sentiment analysis compared to other machine learning algorithms. This might be due to the high complexity of natural language which is difficult to be handled by unsupervised learning methodology. However, using linguistic rules and dealing with the drawbacks of the k-means algorithm, which are low accuracy, group interpretation and instability, has led to a promising unsupervised clustering method.
%^
The proposed method utilizes SWN lexicon to determine the features’ polarity, therefore for languages other than English, an alternative corresponding sentiment lexicon or method for clustering the features is required.

In this paper, we introduce an approach addresses the domain dependency and the intonation cost problems in SA because it is unsupervised labelling-free method. We also introduce two new problems which can be benchmark problems in the sentiment analysis field. The two datasets are Australian AirLines and HomeBuilders reviews datasets, In Airline dataset, the number of reviews for each class (positive and negative) is 750 reviews, whereas HomeBuilders dataset has 1100 reviews for each class. The following are the main contributions of this article:

\begin{itemize}
  \item Introducing a reliable domain independent algorithm through combining contextual analysis and unsupervised ensemble learning.
  \item Two new reviews datasets of Australian AirLines and HomeBuilders collected and tested on along with other datasets. 
  \item New method to address intensifiers and negation using SWN, in addition to considering contrast.
  \item Modifying k-means via using SWN to generate two initial seeds, and discussing a reliable method groups interpretation.
  \item A comparative results discussion is provided.
\end{itemize}

 The organisation of the remainder of the article is as follows: section \ref{RelatedWork} gives a review of the related work. In section \ref{Methodology}, we describe the algorithm and give a background on the related methods. Section \ref{Experiments} presents the experiment data and analysis. In section \ref{Conclusions}, a conclusion is drawn. 

\section{Related work }
\label{RelatedWork}
Sentiment analysis research has taken different research directions such as lexicon-based methods and machine learning which includes several techniques. In the following, important research directions are discussed.

\subsection{Lexicon-based methods}
\label{subsec1}
Lexicon-based methods were the earliest methods suggested for sentiment analysis. These methods are regarded as symbolic approaches because they simply rely on the appearance of documents' terms in a lexicon. Where usually documents are classified by aggregation of sentiment polarity scores from a lexicon. The proposed lexicon methods are different in the lexicon that has been used or generated. Manually generated lexicons such as MaxDiff \citep{kiritchenko2014sentiment}, MPQA \citep{wilson2005recognizing} and General Inquirer \citep{stone1966general} contain comparably less terms and usually a sentiment score or label is associated with each word. Labelling or scouring terms manually is subjected to the annotators’ judgement which can be inconsistent and unequal in terms of accuracy from word to word.

For automatically generated lexicons \citep{klenner2014inducing}, such as SentiWordNet (SWN)\citep{esuli2006sentiwordnet}, the number of terms is usually high and generally no human participation is required. Thus we utilize this lexicon in the proposed method, as it will be demonstrated in section \ref{Methodology}. One of the earliest work was the lexicon-based method by \citet{hatzivassiloglou1997predicting}, where adjectives conjoined by "and" or "but" were used to build a lexicon, then clustering a graph that was produced by using the generated lexicon.

Recent studies \citep{taboada2011lexicon,er2016user} have combined a lexicon-based approach with other techniques such as linguistic rules and neural networks. This is because methods based solely on a lexicon are usually inferior in terms of accuracy due to neglecting changes in the actual sentiment strength when a term appears in different contexts.

\subsection{Contextual analysis methods}
\label{subsec5}
Contextual analysis methods are usually addressing common linguistic structures such as negation and contrast, which also were referred to as sentiment shifters \citep{liu2012sentiment}, sentiment modifiers \citep{muhammad2016contextual}, polarity shifters \citep{xia2016polarity, li2010sentiment} and valence shifter \citep{taboada2011lexicon,polanyi2006contextual}. These linguistic expression forms can be detected using some explicit hints such negation terms. For obtaining higher accuracy, in many studies, the usage of contextual rules is a preceding step which is followed by a machine learning technique. 
In \citep{xia2016polarity}, a rule-based method is proposed to detect text contains a negation and contrast, which was used to train a component classifier of an ensemble method. In addition to that, another component classifier was trained on processed reviews, where the negations have been removed and an antonym dictionary was used to replace the negated terms. The dictionary was built via deploying a weighted log-likelihood ratio (WLLR) algorithm. In order to tackle the ambiguity of the contextual linguistic structures some studies \citep{taboada2011lexicon, polanyi2006contextual,muhammad2016contextual} proposed lexicon and rule-based methods. The idea is to adjust the sentiment polarity scores, which will be extracted from a sentiment lexicon. Score adjustment is suggested because the sentiment score of any term in a lexicon will not precisely reflect the actual sentiment strength of this term when appears within a textual form.  For example "very good"  it is, obviously, invoking a strong positivity, the prior polarity score of the term "good" have been modified by the word "very". 
 %^

We apply an automatic contextual analysis as a first phase to increase the accuracy rate because utilizing clustering solely, which is the second phase, will not yield a good performance. For performing some of the contextual procedures, instead of adjusting terms sores, specialized dictionaries are built and deployed using SWN, which are more effective as been observed experimentally.  
\subsection{Supervised machine learning algorithms}
\label{subsec2}

Most of the proposed machine learning methods for sentiment analysis are supervised. In the initial work, single classic data mining classifiers were usually leveraged such as SVM, ME and NB. A well-known study by \citet {pang2002thumbs}, who conducted experiments using three supervised machine algorithms, NB, ME and SVM, is considered a cornerstone work in this field. As reported by Pang and Lee and many other researchers, SVM usually results in higher accuracy compared to other classic data mining approaches. In the later work more complex supervised learning algorithms were suggested to address the natural language complexity, such as ensemble algorithms \citep {xia2011ensemble, xia2016polarity, wang2015pos} where certain mechanisms can be used to combine results of several classifiers and vector space models, which can increase the accuracy rate.
However most of them were composed of supervised learners whcih need to be trained on labeled data, thus sufferings from the domain dependency problem and usually can not deal effectively with completely unseen data \citep {zhou2015cross}. Therefore, to overcome this problem, our method uses an unsupervised learning paradigm to address SA.

\subsection{Clustering-based approaches}
\label{subsec6} 
Clustering-based approaches to sentiment analysis have been considered in a few research studies. A sentiment analysis approach, which leverages the k-means clustering algorithm, was introduced by \citet{li2012application}. Their solution for k-means instability is to apply a voting mechanism, in which a voting by multiple results of k-means decides the group membership of a document, therefore this approach can also be considered as an ensemble learning method. Using the TF-IDF weighting method with adjective and adverb features effectively increases the accuracy rate by more than 15\%. WordNet has been used to enhance the performance by obtaining the term score, which led to an increase in the accuracy rate. However, the approach relies on a random centroids selection which can affect its stability and performance. The method also relies on experimentally chosen seeds for groups' identification this means the seeds have to be selected each time new data is processed. Thus we propose a nonstochastic centroids selection for clustering, and automatic group identification based on initial centroids and ensemble method. In \citep{ma2013comparison}, a comparative study was conducted on several clustering algorithms and with different weights. They reported that k-means is suitable for balanced datasets. Another comparative work by \citet{ma2013comparison} shows the impact of some weight schemas, and they report that k-means results in higher accuracy on average. The notion of the unsupervised clustering approach can be an effective approach for sentiment analysis if it is reliable and its results are comparably accurate. 

As explored above, in the related work review, There is a domain dependency and data annotation issues when using supervised learning. Therefore this work introduces an unsupervised hybrid approach that combines contextual analysis and clustering classifiers ensemble. The contextual analysis phase is considered because it increases the accuracy rate which makes the method comparably effective. In following section the contextual analysis and ensemble phases will be discussed in detial.

An Automatic Contextual Analysis and Clustering Classifiers Ensemble

\section{An Automatic Contextual Analysis and Clustering Classifiers Ensemble (ACACCE)}
\label{Methodology}

\begin{figure}[H]
\centering
  \includegraphics[scale= 0.3]{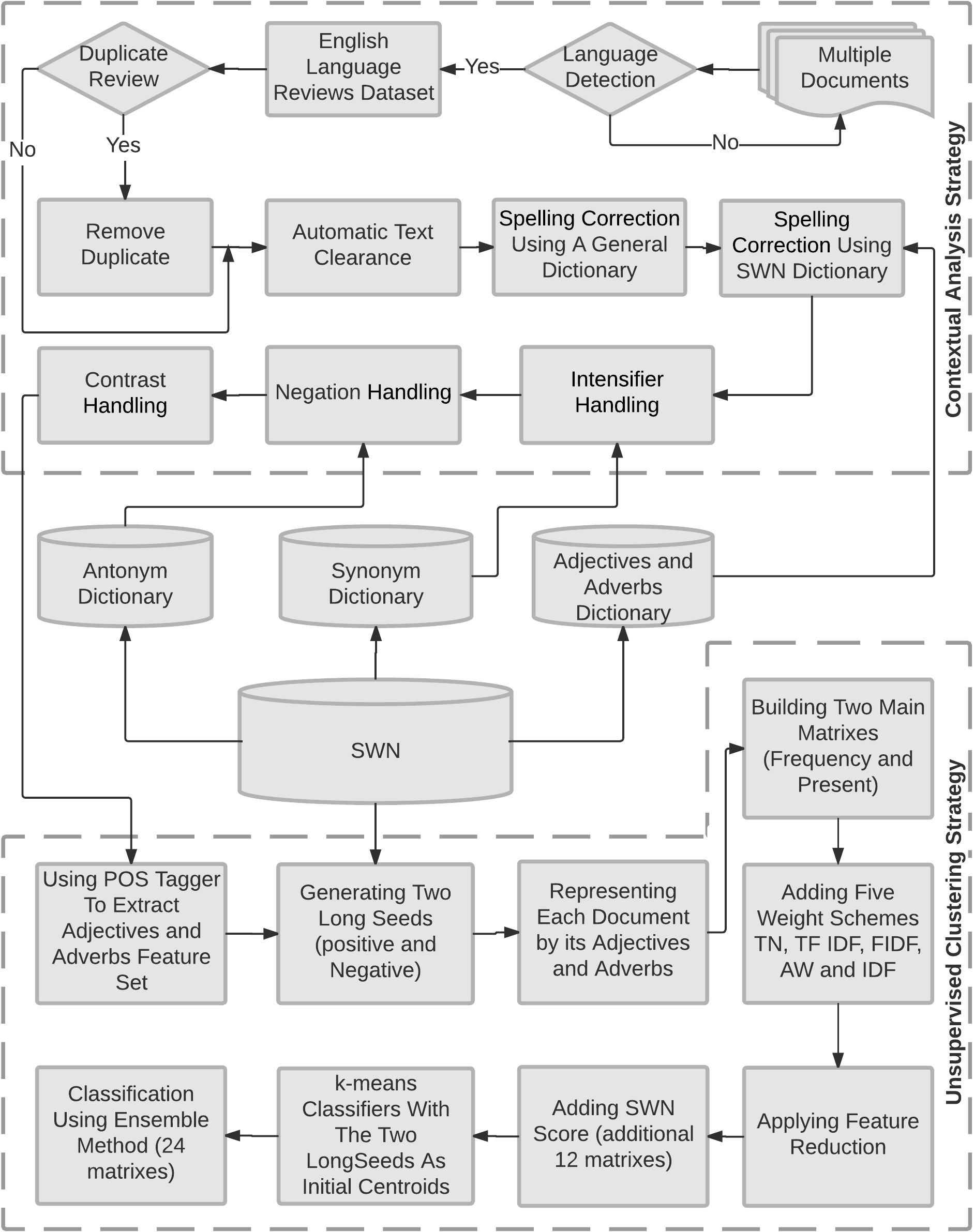}
  \caption{Flow chart of an automatic contextual analysis and  clustering classifiers ensemble (ACACCE)}
  \label{fig:Diagram}
\end{figure}

In this section, we will present the two phases of ACACCE (Figure \ref{fig:Diagram}) to classify products' reviews. The first phase is the data preparation and contextual analysis phase where steps are taken to automatically prepare and clean the text, which are followed by processing common language phenomena, such as intensifiers, negation, and contrast. The second phase is an unsupervised clustering classifiers ensemble where k-means is a base classifier.

\subsection{An Automatic Contextual Analysis}
\label{phase1} 
Contextual analysis is the first phase of ACACCE, which comprises five automatic and consecutive processes for preparing reviews and tackling common language forms. SWN has been effectively utilized to generate specialized dictionaries for some of these processes. The five processes are (1) data preparation; (2) Spelling correction; (3) intensifier handling; (4) negation handling; and (5) contrast handling.

\subsubsection{Data Preparation}
\label{DataPreparation} 

\paragraph{Language Detection} The first step of the algorithm is detecting the language using a detection language tool implemented by Cybozu Labs\footnote{http://labs.cybozu.co.jp/en/}. The library utilises a Naive Bayesian classifier and was reported to achieve over 99\% accuracy for 53 languages. Language detection is considered because we are interested in classifying reviews written in the English language only and it is likely that processed online text will contain reviews that are written in languages other than English.
 
\paragraph{Data Clearance} This step enables automatic data clearance which is significant when processing row online text. The clearance method is role-based and involves removing duplicated reviews and XML tags. It also involves processing each review to separate non-separated tokens and sentences which results in more accurate sentence boundary detection and tokenization process in the following processes.

\subsubsection{Spelling Correction}
\label{SpellChecking} 

Spelling Correction plays an important role in ACACCE because misspelled terms cannot be processed in the following analysis which uses tools such as a POS tagger and dictionaries such as the antonym dictionary. Thus, correcting as many misspelled terms as possible can significantly enhance performance, especially if there is a large number of misspelled adjectives and adverbs because these parts of speech will be a features set of the unsupervised classification. Correcting misspelled words is addressed using two dictionaries, one being a general dictionary after which a specialized dictionary derived from SWN \footnote{http://sentiwordnet.isti.cnr.it/} is used to correct the adjectives and adverbs in the reviews.

\subsubsection{Intensifier Handling} 
\label{Intensifier} 
An intensifier is normally an adverb in a sentence. An intensifier quantifies the strength of an adjective. For instance, in the sentence "the performance was extremely successful", "extremely" is an intensifier, which shifts the sentiment from positive to extremely positive.  
Intensifiers are common and effective sentiment shifters, addressing this type of polarity shifter improves the algorithm's performance. We deal with intensifiers by utilizing a synonym dictionary that has been generated from SWN. The dictionary contains all the adjectives and adverbs in SWN where each synonym pair is chosen to be of the same or close sentiment score regardless of their semantic meaning. We focus on adjectives and adverbs because they will form the feature set for the clustring phase. A predefined list of intensifiers is defined and used in the process to identify the intensifiers. Intensifier in ACACCE is handled by replacing the intensifier with a synonym of the intensified term.
Let $I = \{I_1, I_2,\dots, I_n\}$ $(n > 0)$ be a sequence of intensifiers and $A = \{A_1, A_2, . . ., A_m\}$ $(m > 0)$ be the sequence of adjectives. In the sentence $S = \{w_1, w_2,\dots, w_l\}$ $(l > 0)$ if there exists $k (k >0)$ such that $w_k = I_i$ and $w_{k+1} = A_j (1 \leq i \leq n$ and $1 \leq j \leq m)$, then $I_i$ is an intensifier of $A_j$, and $I_i$ can be replaced by $A'_{j}$ which is a synonym of $A_j$. 
For instance, given the expression "so popular", the word  "so" is replaced with the word  "palmy" which is the synonym of "popular" in the dictionary. In this way, the invoked sentiment, whether it is positive or negative, can be detected by adding synonyms which will be extracted as features for the next learning phase of the algorithm.

\subsubsection{Negation Handling}
\label{Negation} 
Another common explicit language form is the negation. Where terms' polarity in a negative statement can be shifted to the opposite polarity. For instance, the word "didn't" shifts the statement polarity of "I didn't like the movie". To identify negative statements, we use a predefined list of negation terms such as "not" and "never". Then, to process negation, we build an antonym dictionary to replace adjectives and adverbs, that follow negation terms, with their opposite sentiment words. Therefore in the above example, after removing the negation term, the word "like" may change to "hate". A similar approach was suggested in \citep{xia2016polarity}, however, our method differs in that it processes positive/negative adjectives and adverbs only and also, we used SWN to build the dictionary. The dictionary is a list of pairs of polar terms which have been extracted from SWN. The antonym words are antonyms in terms of sentiment strength, regardless of their actual meaning. When using a rule-based method to process negation, the scope affected by the negation term needs to be specified. We tested different scopes and experimentally found that a five-word scope after a negation term is the most effective.

\subsubsection{Contrast Handling}
\label{Contrast} 
Contrast is another commonly used language structure in English. When a sentence contains a contrast term, it will have a clause that concludes author's opinion which will be focused on to determine the sentiment of the sentence. For example, in the sentence "It is a classic feel movie but unfortunately being a cynic", the overall sentiment is expressed in the part that follows the contrast word "but" which is "unfortunately being a cynic". To identify the contrast in a document, a list of pre-defined contrast words, such as  "but" and  "however", has been identified. Then, every sentence in each review which contains a contrast term is processed separately.   Let $S=\{w_1,w_2,\dots w_m\}$  be the sentence which includes contrast terms,  and $C=\{c_1,c_2,\dots c_n\}$ denotes the set of contrast terms. All words $w_j$ of the revoked part will be removed and words in the conclusion part will be kept.

\subsection{Clustering Classifiers Ensemble}
\label{phase2}
Ensemble learning is an effective technique, especially when the targeted data is complex and can be represented in many forms. Although the ensemble learning imposes a higher complexity compared to a singleton classifier, it is capable of generating a model of high diversity, which enhances the power of generalization. Therefore we have used this technique with several vector space models, each model represents the dataset in a unique weight scheme. The ensemble component classifier is k-means which is well known unsupervised clustering algorithm. All the documents are represented by their sentiment expressing words such as adjectives and adverbs \citep{benamara2007sentiment}.

	\subsubsection{k-means algorithm}
	\label{k-means} 
	
	The component classifier of the proposed ensemble method is the k-means algorithm. It is a statistical and conventional clustering mean with hard boundaries in which the produced clusters are of unshared instances. It is a simple, flat, hard and polythetic clustering algorithm, with a predefined number of clusters. Several researchers have contributed to the design of the algorithm for different disciplines.

The algorithm is suitable for our experiments because (1) k-means is unsupervised clustering algorithm therefore it is suitable for a domain-independent method. (2) k-means will always converge with a low number of iterations \citep{arthur2007k}, which we also observe experimentally (Refer to Table \ref{table:componentClassifiers1} and \ref{table:componentClassifiers2}). The low iterations number is also a result of a proper centroids selection. (3) Although predefining clusters number and hard clustering can be considered drawbacks of k-means, it is adequate for the method because ACACCE preduces only two positive and negative clusters, and by using k-means we can pre-assign the numbers of clusters. (4) k-means instability is addressed via non-random initial centroid selection, which also enhances its accuracy.
 
The default k-means is initiated by selecting $k$ random centroids (vectors) from a given dataset \citep{wu2008top}. Firstly, the centroids are randomly selected after which each data point is assigned to its closest centroid via a similarity measurement such as Cosine distance or Euclidean distance or other appropriate measurement methods can be applied. The next step is setting the average of the clustered points in each group as the new centroid for the corresponding cluster. Then, by iteratively recalculating the closest distances, the cluster means and setting new centroids to the obtained groups, the convergence condition is obtained when no new centroids are found, or in other words, no point is reassigned to a cluster.

The performance of k-means is highly influenced by (1) the initial centroid selection; (2) data representation; and (3) distance measurement, we have chosen Cosine distance because prior experiments have shown that Cosine distance leads to more accurate results. Below, we describe our attempt to enhance the performance of ACACCE using k-means as a based classifier through mainly focusing on first centroids selection using SWN, and data representation.

	\subsubsection{SentiWordNet}
	\label{subsubsec2}
The specialized sentiment lexicon SentiWordNet 1.0 (SWN) \citep{esuli2006sentiwordnet} is an automatically generated lexicon in which three scores (positive, negative and objective) are assigned to each synset from WordNet. We utilize SWN to perform a contextual analysis and determine the features of the polarities which form the initial k-means centroids. 
	
\begin{figure}[h]
\centering
  \includegraphics[scale= 0.3]{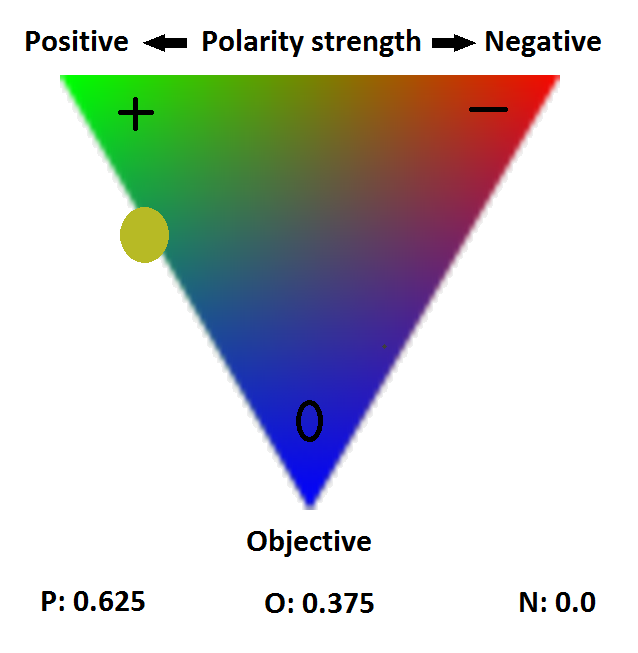}
  \caption{Example of SentiWordNet 3.0 online graphical representation of first sense of the word ‘faithful’ as an adjective.}
  \label{fig:SWN}
\end{figure}	

	The scores measure the strength of the terms’ polarity by assigning a value for each of the three classes, where the total of these values is equal to 1.0, and each class has a partial value based on the strength of the three invoked sentiments. A committee of eight ternary semi-supervised classifiers was utilized to build this lexicon. In this work, the enhanced version SentiWordNet 3.0 \citep{baccianella2010sentiwordnet} is used which is based on WordNet 3.0, the updated version of WordNet, where, in addition to the committee classifier, Random walk was used to enhance the scores. An improvement of over 19\% was  reported \citep{baccianella2010sentiwordnet} when using the updated version. 
	
The SentiWordNet’s scores are generated based on WordNet’s synsets. Therefore, the same word in SentiWordNet can have different scores, for it may appear in several synsets. Thus, the average of the synsets’ scores for each term is used instead of the synset polarity values which requires a text ambiguity analysis approach, which is another research direction that will not be covered in this work. The final score of a term is expressed in equation (\ref{equ:SWN}).		
		%equation
		\begin{equation}
		\label{equ:SWN}
		finalScore=pos-neg
		\end{equation}
		
	where $pos$ is the average of the positive scores of each term, and $neg$ is the average of the negative scores of each term.
	\begin{algorithm}[H]
	\textbf{INPUT:} A SentiWordNet Lexicon $ L $ contains all terms $ w_m $ \\
	\textbf{OUTPUT:} A set $ U$ of unique adjectives and adverbs $ u $ with averaged scores
	\begin{algorithmic}[1]
	\ForAll{terms $ w_m $ in $ L $}
		\ForAll {unique adjectives and adverbs  $ w_j \in L$ where $ L $ is the SentiWordNet Lexicon}
			\If { $ t_j==a$ OR $t_j==r, t_j$ is part of speech tag,  $a$ and $r$ donate adjective and adverb respectively}
				\State $vPos_j=\frac{1}{n} \sum_{(i=1)}^{n}PositiveScore_i$ , where $n$ is the synsets number
				\State $vNeg_j=\frac{1}{n} \sum_{(i=1)}^{n}NegativeScore_i$ , where $n$ is the synsets number
				\State Assign $finalScore_j= vPos_j- vNag_j$ to $u_j$
				\State Add $(u_j,finalScore_j)$ to $U$
			\EndIf
		\EndFor
	\EndFor
	\end{algorithmic}
	\caption{Extracting a set of unique adjectives and adverbs from SentiWordNet with averaged scores }
	\label{algo:}
	\end{algorithm}
	
	\subsubsection{Initial centroids}
	\label{subsubsec3}
The initial selection of k-means centroids is an important factor in forming the final clusters, hence the process and outcome of clustering to a certain degree depends on the first iteration where the initial centroids are selected \citep{wu2008top}. The selection of first centroids seems to be the main factor that affects the iterations number and the convergence of the algorithm.
%^

A random selection, in default k-means, can result in poor performance because in the process of binary classification, for example, these two randomly selected points can be of the same class, which will lead to inaccurate clustering based on similarity. Some suggestions were introduced to address this problem such as k-means++ \citep{arthur2007k} which selects distanced random points where the probability of choosing each of these points is proportional to its overall potential contribution. However, a selection of initial noninformative points or outlier points, which are uncorrelated and dissimilar from any other documents, is another reason that can reduce classification accuracy, in spite of selecting dissimilar centroids. Other studies suggest genetic algorithms to address this issue \citep{laszlo2007genetic,babu1993near}. Operating the algorithm several times is another suggestion to overcome the drawback of the random selection of the first centroids. In \citep{li2012application}, several results of k-means runs were combined using a voting mechanism, however, their method still based on a stochastic initialization and it does not completely eliminate the instability problem.

In our approach, ACACCE, we use SWN to automatically generate two polar seeds (positive and negative), then insert them into the dataset to be assigned as initial centroids of the k-means algorithm.  These two polar centroids are guaranteed to be of different classes (ie. distanced points) that are always informative and can correlate with most of the documents in the processed data. Nonrandom initialization eliminates the instability problem because k-means will always produce the same clusters when operating on same dataset. The positive and negative seeds are automatically extracted from the feature set which is the set of adjectives and adverbs in all processed documents. SWN is utilized to split the feature set into three sets, positive, negative and neutral. This is implemented by matching each feature in the feature set against the SWN lexicon score. Then the positive feature set is assigned as a positive initial centroid and the negative feature set is assigned as a negative initial centroid. This is an effective and efficient solution to the instability of k-means, because the process will start with informative centroids that contain all the negative/positive features. 

	\begin{algorithm}[H]
	\textbf{INPUT:} A set of feature $ F $\\
	\textbf{OUTPUT:} Insert the positive $PosD$ and the negative $NegD$ seeds into the corpus $D$
	\begin{algorithmic}[1]
	\ForAll {$f_{i(1 \leq i \leq size(F))}\in F$, where $F$ is the feature set.}
		\State Match $f_i$ against $finalScore_j$ of  $u_j \in U$, $U$ is the extracted set of adjectives and adverb from SentiWordNet
		\If {$ finalScore_j > 0$ }
		\State Add $f_i$ to $PosD$ , $PosD$ is the positive document
		\ElsIf
			{$finalScore_j < 0$}
			\State Add $f_i$ to $NegD$, $NegD$ is the negative document
		\EndIf
	\EndFor
	\State 	Insert $PosD$ and $NegD$ into $D$, $D$ is the corpus
	\end{algorithmic}
	\caption{Insert the two seeds}
	\label{algo:}
	\end{algorithm}

	\subsubsection{Vector space models}
	\label{subsubsec5}
	The vector space model is a commonly used representation in text processing, where terms are features and documents are observations. A variety of matrix representations has been examined to obtain the most accurate results. With the proposed system, it is possible to experiment with 24 vector space models which are different representations of about 2000 documents of each dataset in a comparably short time. This is mainly because it is an unsupervised system, and the use of k-means with nonstochastic centroids selection results in a reduction in the computational complexity of the method. To build various VSMs, two matrixes are generated, namely the presence matrix and the frequency matrix. In the presence matrix, each document is represented by a binary fixed length row, and value of 1 represents the presence of particular features in a document. The frequency matrix, which is commonly called a bag of words in the literature, represents each document as a fixed length row in the vector space and each value is the count of the feature’s occurrence in this document. For both matrixes, the following weights were used in the experiments and the results are shown in Table \ref{table:componentClassifiers1} and \ref{table:componentClassifiers2}. In addition to those weights, the VSM number is increased by adding SWN scores to each matrix (Figer \ref{fig:ensamble}).

\paragraph{Term normalization (TN) \citep{croft2010search}}		
TN measures the importance of a term in a particular document, where the numerator is a word’s count and the denominator is the length of a document where the word occurs. It expresses the importance of the word, taking into consideration the differences in the documents’ length. It seems to be a reasonable method in dealing with documents of unbalanced length, as in the set of movie reviews, where, for example, the shortest document contains only 17 words and the longest consists of over 2500 words. Equation (\ref{equ:TN}) is the mathematical expression of term normalization.	
		%equation
		\begin{equation}
		\label{equ:TN}
		tf_{i,j}=\frac{t_{i,j}}{l_j}
		\end{equation}
where $t_i$ is the frequency of term $i$, $l_i$ is the length of document $j$ where term i is occurred.

\paragraph{Inverse document frequency (IDF)}
IDF  is usually used as a part of another weighting method, along with the measurement of term importance or frequency in a document.  It measures the importance of a term in a given corpus, regardless of the term’s importance in a particular document. As empirically observed, using IDF can be more effective in some circumstances than combining it with another term’s importance measure. Equation (\ref{equ:IDF}) is the mathematical expression of IDF.
		%equation
		\begin{equation}
		\label{equ:IDF}
		{idf}_i=\log\left(\frac{D}{{df}_i}\right)	
	\end{equation}
	
	Let $D$ be the number of all the documents and $df_i$ is the number of documents where term $i$ occurred.
		
\paragraph{Term frequency–inverse document frequency (TF-IDF) \citep{croft2010search}}
TF-IDF is a plausible and commonly used scouring mechanism in text mining tasks. It measures the importance of a particular word not only in the document, via term frequency (term normalization) as previously discussed, but also in the corpus via inverse document frequency. TF-IDF is proportional to the term frequency value and offsets the inverse document frequency value. It is expressed mathematically by equation (\ref{equ:TF-IDF}). 
		%equation
		\begin{equation}
		\label{equ:TF-IDF}
		{tf-idf}={tf}_{i,j}*{idf}_{i}
		\end{equation}		
		where $tf_{ij}$ is the term normalization of term $i$, and $idf_i$ is the inverse document frequency of term $i$.
\paragraph{Weight frequency-inverse document frequency (WF-IDF) \citep{manning2008introduction}}
WFIDF is another common weight mechanism that has been proposed to improve the accuracy of text mining systems. It is a proposed solution to the drawback of using term frequency which is the assumption that the count of appearances of a term in a document is equal to the significance of a single occurrence. Equation (\ref{equ:WF-IDF}) is the mathematical expression of WF-IDF. 
		%equation
		\begin{equation}
		\label{equ:WF-IDF}
		{wf-idf}=\begin{cases}
		1+\log{tf}_{i,j}.{idf}_i,&\text{ if ${tf}_{i,j}>0$ }\\
		0, &\text {Otherwise} 
		\end{cases}
		\end{equation}
		where $tf_{ij} $is the term normalization of term $i$, and $idf_i$ is the inverse document frequency of term $i$.
\paragraph{Average of weights (AW)}
The average of the two weights, TFIDF and WFTDF, is calculated and the obtained results can be more accurate than using a single weight scoring method (Table\ref{table:componentClassifiers1}). Equation (\ref{equ:AW}) is the mathematical expression of the averaged weight of term $i$ in document $j$.		
		%equation
		\begin{equation}
		\label{equ:AW}
		AW=\frac{({tf-idf}_{i,j})+({wf-idf}_{i,j})}{2}
		\end{equation}
		\begin{algorithm}[H]
	\textbf{INPUT:} A corpus  $ D $ \\ 
	\textbf{OUTPUT:} A set of matrix files $M_n$
	\begin{algorithmic}[1]
	\State 	Create $M_n$ empty matrix files, $n$ is $24$ matrixes
	\ForAll {document  $d_j \in D$} \label{marker}
		\State Create a presence vector $vp_j$, and frequency vector $ vf_j $
		\State Add $vf_j$ to $M_1$
		\State Add $vp_j$ to $M_2$
	\EndFor
	\ForAll{vectors $vf_j  \in M_1$ and  $vp_j \in  M_2 $}	
		\ForAll{feature $f_i$ }
	 	 \State $ f_i*weigth_i$, $weight_i$ denotes $TF,IDF,TF-IDF,WF-IDF $ and $ AW $
		\EndFor
		\State	Add the new vector $ v_{j=\{1, \dots ,10\}} $ to $ M_r (2>r>13)$
	\EndFor
	\State Remove the neutral features
	\State Add $finalScore_j$ to all $12$ matrix files, another $12$ matrixes files of  $M_n$ will be filled  
	\end{algorithmic}
	\caption{Constructing the vector space models}
	\label{algo:}
	\end{algorithm}	
		
\subsubsection{Neutral term and feature reduction }
\label{subsubsec6}
Neutral terms can be considered as redundant features for our experiment because we are interested in two classes only, positive and negative, and it is assumed that no sentiment polarity is likely to be expressed by the neutral features. Therefore, feature reduction can be conducted by eliminating the neutral terms. Careful consideration should be given, before using other feature selection methods after removing the neutral features as it may lead to the inaccurate clustering of short documents in the high sparsity vector space.

\subsubsection{Cluster interpretations }
\label{ClusterInterpretations}
The k-means algorithm requires an interpreting strategy when processing real-world data because no labels will be provided to interpret the acquired groups. In \citep{li2012application}, group polarity is judged based on the distribution of solid polarity documents in the clusters, and its solidity has been proven experimentally by observing 100 clustering results, where 22 documents were always correctly classified. Despite the low possibility of incorrectly classifying these seeds, which is $10^{-z}$ where $z$ is the number of positive/negative seeds, this low possibility will probably be altered if there is a modification to the dataset size or if another dataset is used.  
In ACACCE, the two seeds that have been used as initial centroids can indicate the clusters’ polarity, and the assumption is that a positive cluster is where the positive seed appears and a negative cluster is where the negative seed appears. The classification of these documents can be described as a flat classification owing to the polarity of all the positive/negative features that form each seed. Thus, even weak classifiers can easily assign the seeds correctly, as observed in the experiments. In order to assess this way of interpretation and test the reliability of using the two solid polarity documents for groups judgement, we discuss two possibilities, one when both seeds are misclassified and second when one of them is misclassified.
To examine the possibility if a classifier labels the positive seed as a negative instance, and the negative seed as a positive instance, we compared our method with \citet{li2012application}'s method, which is based on the confusion matrix. In the confusion matrix (refer to Table \ref{table:ConfusionMatrix}), where $a, b, c,$ and $d$ are the number of documents therefore, in \citep{li2012application}, if $(b + c) > (a + d)$, Cluster 1 is considered positive and Cluster 2 is negative, otherwise vice versa. The comparison between the interpretation using confusing matrix and interpretation using seeds method shows that throughout all the experiments, no contradiction between the two methods. The interpretation is mostly similar, except when the two seeds appear together in one group, which is the second possibility where one seed is misclassified, here the ACACCE method gives no interpretation and neither group is determined as positive or negative. Thus, this classifier's results will not be considered in the ensemble. To this end, utilizing the seeds can be considered a reliable indication of the groups identification because a component classifier always will either correctly classify the two seeds or misclassify one of them, in which the classifier will be neglected. 

\subsubsection{Ensemble Learning}
\label{EnsembleLearning}

\begin{figure}[h]
\centering
  \includegraphics[scale= 0.3]{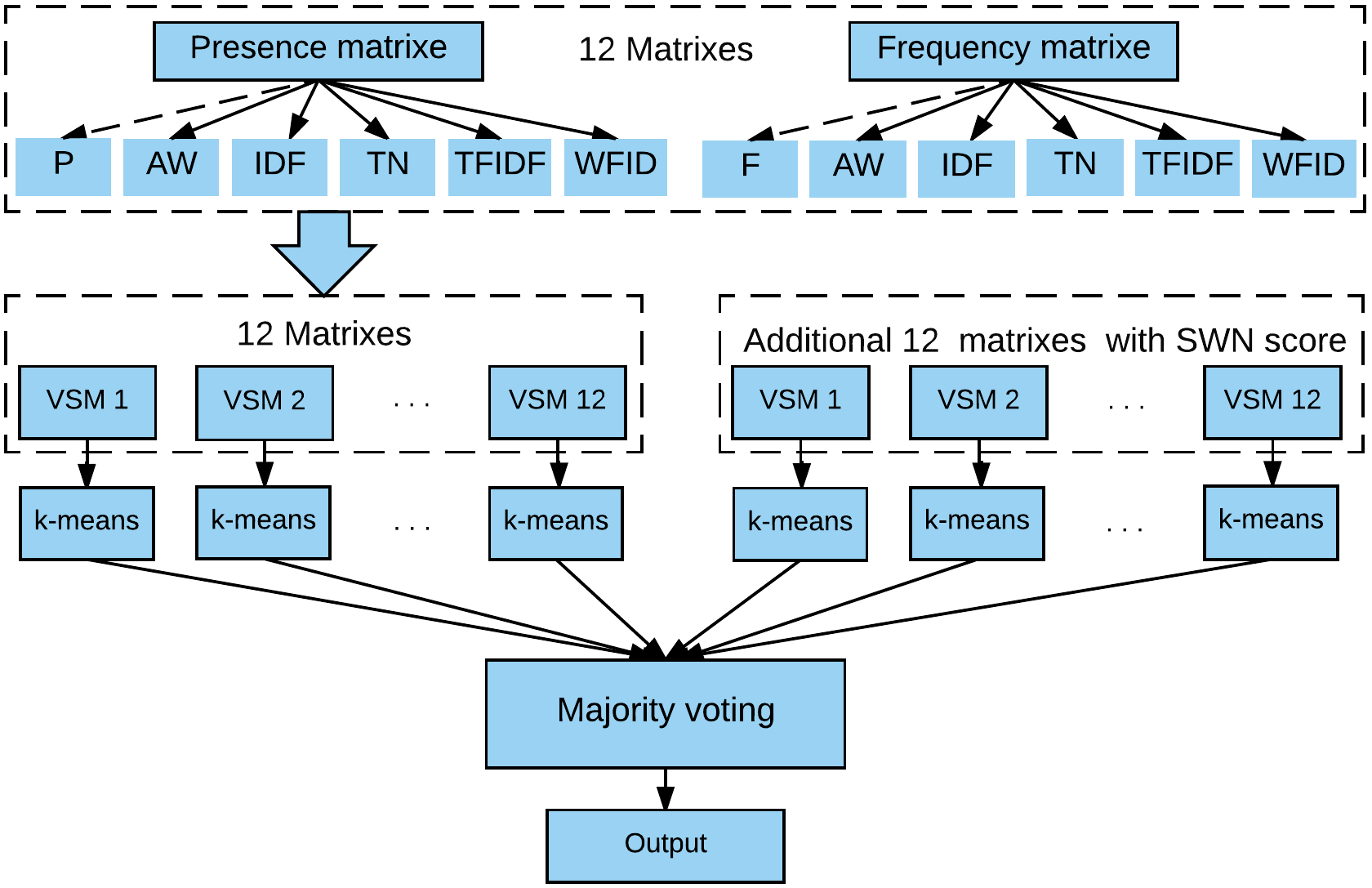}
  \caption{Ensamble method}
  \label{fig:ensamble}
\end{figure}

Ensemble learning (Figure \ref{fig:ensamble}) is a combination of several classifiers to achieve higher accuracy. It can combine learners of the same type, for example, bagging and boosting ensemble methods \citep{whitehead2010sentiment,wang2014sentiment}. It can also be an ensemble of different types of classifiers \citep{da2014tweet}. The ensemble algorithms that have been proposed for sentiment analysis are mostly supervised algorithms \citep{wang2015pos,tsutsumi2007movie,li2012heterogeneous,fersini2016expressive}. They differ in the learning and the feature selection stage of the base classifiers and in its base classifier combination methods. The idea is that ensemble can be more accurate compared to a single classifier if the component classifiers are diverse and accurate \citep{hansen1990neural}. An accurate classifier, also referred to as a weak classifier by \citet{schapire1990strength}, according to \citep{dietterichensemble,schapire1990strength} is a classifier which its performance is better than random guessing. An ensemble method often enhances performance because its outcome is a result of base learners’ results being collected and combined in a certain way, such as voting or weighting. As a result, complex problems can be solved, even by a combination of weak classifiers. It can also solve the overfitting problem, avoiding potential computational failure such as stacking in local optima and solving complex problems which might be too difficult to solve using a single classifier \citep{dietterichensemble}. These advantages motivate us to examine the effect of an ensemble method by applying majority voting on the results of the k-means classifier, with pre-specified initial centroids, on different vector space models. The component classifiers are insured to be diverse by using different weight schemes, and also their accuracy is enhanced compared to random guessing by using initial seeds as centroids of k-means. More importantly, in ACACCE, assembling is significant for the groups identification. The chance of inaccurately classifying the initial centroids in ensemble learning is extremely low because most of the classifiers will be able to allocate the seeds correctly, and this seed allocation can be considered a very strong indication of the group meaning. To enhance accuracy, and because a few of the weak learners, as previously mentioned, may misclassify one of the two seeds, these classifiers’ results can be ignored when both seeds appear in the same group.

\subsubsection{Ensemble of clustering classifiers algorithm}
\label{subsubsec9}
The ensemble algorithm is as follows:
		\begin{algorithm}[H]
		\textbf{INPUT:} A corpus $D$ of $m$ number of documents $\{d_1,d_2, \dots ,d_m\} $   \\
		\textbf{OUTPUT:} Assign a Positive OR Negative label For each document $d_i \in D$, $(i=1, 2,\dots , m)$\\		
		 Pre-processing:
		\begin{algorithmic}[1]
		 
		\ForAll{document $ d_j   \in D$}
			\ForAll {each word $w_i \in d_j$}
			\State 	Tag $w_i$ with part of speech tagging $t_j$
			\If {$ t_j==a$ OR $t_j==r $, $a$ and $r$ donate adjective and adverb respectively}
				\State Keep $w_i$
				\State 	Add $w_i$ to $F$, $F$ is the features set 
			\Else
				 \State Remove $w_i$ 
			\EndIf
			\EndFor
		\EndFor
		
Clustering: 
		\State Set the clusters number $K=2$
		\ForAll{matrix files $M_i$, $(i=1,2, \dots ,n)$, }
			\State 	Initialize positive seed $PosD$ and negative seed $NegD$ as first centroids 
			\State 	Cluster $M_i$  into two clusters $G_1$  and $G_2$  by using k-means classifier $H_i$, with cosine similarity
				\If {$PosD \in G_1$ and $NegD \in G_2$}
					\State $H_i$ classifier is accurate enough
					\State 	$G_1$ is the positive cluster, $G_2$ is the negative cluster
				\ElsIf 
					 {$PosD \in G_2$ and $NegD \in G_1$}
					\State $H_i$ classifier is accurate enough
					\State 	$G_2$ is the positive cluster, $G_1$ is the negative cluster
				\Else 
					\State $H_i$ classifier is NOT accurate
			\EndIf
		\EndFor	
		
%\algstore{bkbreak}
%\caption{Pre-processing and the ensemble of clustering classifiers algorithm}
%\end{algorithmic}
%	\end{algorithm}
%	\begin{algorithm}[H]
%	 Voting:
%	\begin{algorithmic}[1]
%		\algrestore{bkbreak}
%\ForAll{$d_j \in D$, $D$ is the corpus}
%	\ForAll{result $ R_i $  of $ H_i $}
%	\If {$ H_i $ classifier is accurate enough}
%		\If { $ \sum(d_j (R_i )=positive \geq \sum(d_j (R_i )=negative ) $}
%		\State 	$d_j=positive$
%	\Else
%		\State $d_j=negative$
%	\EndIf
%	\EndIf
%	\EndFor
%\EndFor
%\end{algorithmic}
%	\label{algo:}
%	\end{algorithm}
%%%%%%%%%%%%%%%%%%%%%%%%%%%%%%%%%%%%%%%%%%%%%%%%%%%%%%%%%%%		
		 Voting:
		 
		\ForAll{$d_j \in D$, $D$ is the corpus}
			\ForAll{result $ R_i $  of $ H_i $}
				\If {$ H_i $ classifier is accurate enough}
					\If { $ \sum(d_j (R_i )=positive \geq \sum(d_j (R_i )=negative ) $}
						\State 	$d_j=positive$
					\Else
						\State $d_j=negative$
					\EndIf
				\EndIf
			\EndFor
		\EndFor
		\caption{Pre-processing and the ensemble of clustering classifiers algorithm}
		\end{algorithmic}
		\end{algorithm}
%%%%%%%%%%%%%%%%%%%%%%%%%%%%%%%%%%%%%%%%%%%%%%%%%%%%%%%%%%%%%%%%%%%%%
			
The idea of combining several VCMs not only leads to more reliable ensemble learning, it also has more flexibility because a future enhancement can be made by using additional weight schemas or another component classifier that is suitable for large data analysis. However, extending the ensemble approach will increase the computational complexity; therefore, another component learner should be carefully selected. 

\subsubsection{Computational complexity analysis}
\label{ComputationalComplexity}
If the complexity of k-means is $O(g(nkt))$ where $n$ is the dataset instances, $k$ is clusters number, and $t$ is iterations number, then the computational complexity of ACACCE is $O(mg(nkt))$. Ensemble methods’ complexity is mostly linear with respect to the number of component classifiers $m$, and it basically depends on the complexity of the base learner. In addition, the computational cost of cosine distance which is the similarity measurement of the base classifiers depends on the vector length. Therefore feature reduction can slightly improve the performance.

\section{Experiments and analysis}
\label{Experiments}
In order to evaluate the method, we conduct experiments on differnt reviews datasets (refer to Table \ref{table:Datasets}). For evaluation, usually, when using machine learning algorithms, an experimental dataset is divided into training and testing portions. In ACACCE, the entire dataset is used for evaluation because it is an unsupervised method. The positive/negative labels that are attached to each document are used to construct a confusion matrix. 
 
As we are interested in both negative and positive classes, the evaluation is done by calculating the accuracy \citep{manning2008introduction,hubert1985comparing}.  Equation (\ref{equ:accuracy}) is a mathematical expression for calculating accuarcy based on the confusion matrix and the seeds’ position. In addition to accuracy, we also calculate precision, recall and F-measure \citep{manning2008introduction}.

\begin{table*}[h]
\centering 
\caption{Confusion matrix }
\begin{tabular}{ccc} 
\hline
Actual negative & Actual positive\\\hline
$ a $&$ b $ & Positive\\ 
$ c $&$ d $ & Negative\\ \hline
\end{tabular} 
\label{table:ConfusionMatrix} 
\end{table*}
%equation

\begin{equation}
\label{equ:accuracy}
accuracy= \frac{a+b}{a+b+c+d} 
\end{equation}

ACACCE is implemented with java 8 and NetBeans IDE 8.0.2. The experiments were conducted on a Dell machine with an 3.40 GHz Intel Core I7 CPU and 16GB RAM, running Windows7 Enterprise. For compersion with other machine learning classifiers we used Weka 3.8.1.    

\subsection{Datasets}	
\label{subsec1}
The across domain performance of ACACCE is evaluated by collecting and experimenting on two datasets which are Australian Airlines and HomeBuilders reviews datasets. We also conduct experiments on movie dataset and multi-domain datasets \citep{blitzer2007biographies} (refer to Table \ref{table:Datasets}).
 
\subsubsection{Airlines and HomeBuilders datasets}
\label{}
A publicly available online reviews is collected from www.productreview.com.au which is an Australia consumer opinion website. Where each review is associated with one of five rating categories (excellent, good, ok, bad and terrible). This enables us to select reviews with excellent and good rating as positive instances, whereas reviews with bad and terrible rating as negative instances. 

\paragraph{Airlines dataset} For constructing this dataset, 1500 reviews on four Australian airlines are randomly collected. Those reviews were written between September 2006 and January 2017. 
%tiger-airways , jetstar, qantas, virgin-blue
\paragraph{HomeBuilders dataset} The collected reviews of this dataset are on 14 home builders companies in Australia, which were written between January 2009 and January 2017. 
%metricon, clarendon, simonds, porter-davis-homes, hotondo-homes, boutique-homes, 8-homes, Burbank, 
%homebuyers-center, beechwood-homes, dixon-homes, watersun-homes, 
%plantation-homes, kurmond-homes. 

\subsubsection{Movie and multi-domain datasets}
\label{subsec1} 

The movie review dataset by \citep{pang2004sentimental}, which is the enhanced version of \citet{pang2002thumbs}'s dataset, is a well-known dataset in the field of sentiment analysis and has been used in many research studies. It is widely believed that movie reviews are difficult documents to classify compared to other product reviews \citep{chaovalit2005movie,wollmer2013youtube}. This is because many aspects are likely to be discussed and different polarities can be invoked. The wide variety of movies can complicate this task even further because of the number of subjects being discussed in the reviews, such as the plot of the movie, the actors, and the movie’s location. It is also likely to contain unbalanced samples of different lengths, which can also cause difficulties in analyzing short documents.

The multi-domain dataset \citep{blitzer2007biographies} is a benchmark dataset which was constructed by \citet{blitzer2007biographies} using reviews on different products taken from Amazon.com. Four domains' review sets have been used in the experiments. The datasets have the same balanced composition which is 1000 positive documents and 1000 negative documents, except Baby products reviews where the review number is 900 for both classes. Each review in both datasets was automatically labeled using the rating information associated with each document, which is provided by the authors.
\begin{table*}[h]
\centering 
\caption{Datasets Table}
\begin{tabular}{|cccc|} 
\hline
 &\multicolumn{2}{c}{Number of samples}&\\
Datasets & Positive & Negative  & Sources \\
 \hline
AirLines &	750 & 750 & http://www.productreview.com.au\\
HomeBuilders &	1100 & 1100 & \\
\hline
Movie \citep{pang2004sentimental} &	1000 & 1000 & http://www.imdb.com\\
\hline
Kitchen \citep{blitzer2007biographies}	& 1000 & 1000 & \\
Apparel \citep{blitzer2007biographies}	& 1000 & 1000 & https://www.amazon.com\\
Toys\&Games \citep{blitzer2007biographies}&	1000 & 1000	& \\
Baby \citep{blitzer2007biographies} & 900 & 900 & \\
\hline
\end{tabular} 
\label{table:Datasets} 
\end{table*}
\subsection{First phase of ACACCE}
\begin{figure}[]
\centering
  \includegraphics[scale= 0.9]{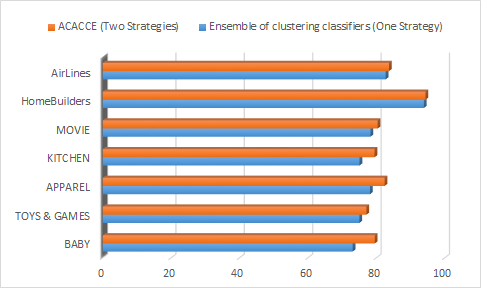}
  \caption{The effect of the first phase on accuracy}
  \label{fig:FirstPhase1}
\end{figure}

The following is a detailed analysis of each procedure of the first phase. Figure (\ref{fig:FirstPhase1}) shows the effect of the contextual analysis phase on accuracy where accuracy rate has increased by an average of 3.01 percent when applying the contextual analysis procedures. 

\begin{figure}[]
\centering
  \includegraphics[]{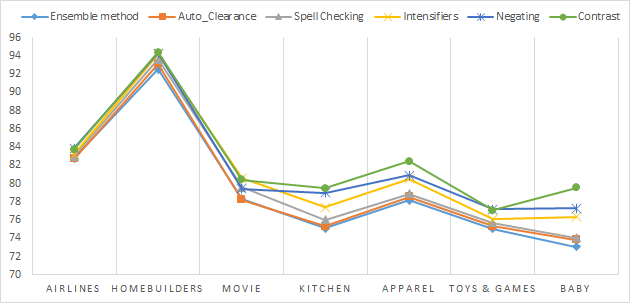}
  \caption{The effect of each procedure of the first phase on accuracy}
  \label{fig:FirstPhase2}
\end{figure}

\textit{Data Preparation:} Figure \ref{fig:FirstPhase2} shows a slight enhancement in accuracy when preparing the data compared to applying the ensemble method to raw data. This step is significant for the following procedures and also for the second phase because it enhances the process of tokenization and sentence boundary detection.  

\textit{Spelling Correction:} Spelling correction using dictionaries positively affects the result because it assists processing and extracting as many adjectives and adverbs as possible in the following steps. When processing raw on-line text, there is a need for data preparation and spelling correction because it is very likely the text will contain misspelled terms. 
  
\textit{Intensifier Handling:} An improvement is noticed when processing the intensifiers, which is due to the strong sentiment intensifying caused by these terms and also to the common use of intensifiers. 

\textit{Negation Handling:} is a common form of language structure which results in strong polarity shifting. As shown in Figure \ref{fig:FirstPhase2} processing negation increased the accuracy in four datasets.

\textit{Contrast Handling:} is the last procedure of the first phase and addresses the contrasts, resulting in a considerable enhancement in two datasets (Apparel and Baby), and a slight enhancement in the other datasets.

As a preceding stage, the contextual analysis procedures improve the outcome of ACACCE (Figures \ref{fig:FirstPhase1} and \ref{fig:FirstPhase2}).

\subsection{Second phase of ACACCE}
\label{subsec3} 
The experiments were conducted by obtaining a high-dimensional matrixes of all adjectives and adverbs as features. For extracting adjectives and adverbs we use Stanford part-of-speech tagger \citep{toutanova2000enriching}. A matrix is representing all documents of each dataset in a vector space model, where each document is a vector in the vector space. This model was proposed for the information retrieval system \citep{salton1971smart}. In this representation of corpus, the order of terms in a document is ignored and the sparsity of the obtained matrix is very high. 

\begin{figure}[]
\centering
  \includegraphics[scale= 0.9]{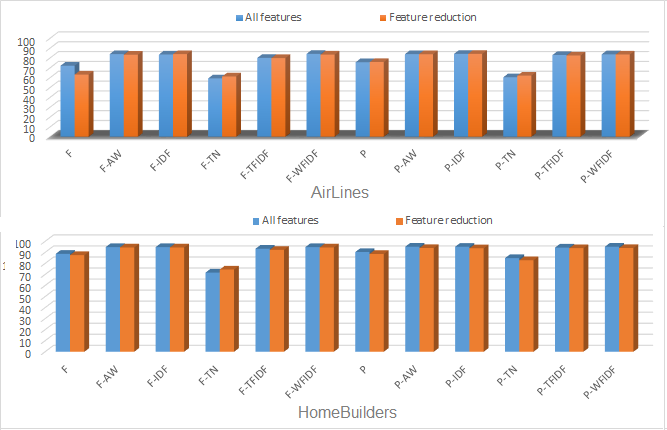}
  \caption{Feature reduction effect on the accuracy}
  \label{fig:FeaturesReduction}
\end{figure}

\paragraph{Vector space models:} The experiments' results of the component classifiers of ACACCE on Airlines and HomeBuilders datasets using five weighting schemes are shown in Tables \ref{table:componentClassifiers1} and  \ref{table:componentClassifiers2}. The first two matrixes that have been tested are the presence matrix and the frequency matrix. Frequency matrix mostly is inferior compared to presence matrix in term of accuracy, the difference between these matrixes' results decreases significantly when the weight schemas are used. One of those weights is TN which always leads to a lower accuracy, probably because it measures the term importance regardless of its importance to the entire corpus. The effect of the terms' weights in the entire corpus becoming clearer when we used the IDF, where the term weight in a particular document is neglected. IDF enhances performance as shown in Tables \ref{table:componentClassifiers1} and  \ref{table:componentClassifiers2}. Using the standard weights TF-IDF and WF-IDF with presence and frequency matrixes significantly enhances the base classifier accuracy by averages over 12\% and 4\% when experimenting on Airlines and HomeBuilders datasets respectively. 
\begin{table*}[h]
	\small
	\caption{Results of the component classifiers on the Airlines using five weighting schemes}
	\begin{adjustbox}{center}
	\label{Results}
	\begin{tabular}{c c c c c c c } 
	\hline
	Matrixes & Accuracy & Precision & Recall & F-measure & Iterations & Time in seconds\\
	\hline
Frequency & 63.16 & 61.22 & 71.9 & 66.14 & 13 &   1\\
Frequency-AW & 83.34 & 87.33 & 78.03 & 82.42 & 19 &   1\\
Frequency-IDF & 83.94 & 87.61 & 79.09 & 83.14 & 12 &   1\\
Frequency-TN & 61.29 & 58.69 & 76.43 & 66.4 & 13 &   1\\
Frequency-TFIDF & 79.95 & 84.19 & 73.77 & 78.64 & 13 &   0\\
Frequency-WFIDF & 83.48 & 87.04 & 78.7 & 82.66 & 12 &   1\\
Presence & 75.95 & 79.55 & 69.91 & 74.42 & 17 &   0\\
Presence-AW & 83.88 & 84.63 & 82.82 & 83.71 & 11 &  0\\
Presence-IDF & 84.21 & 85.99 & 81.76 & 83.82 & 9 &   1\\
Presence-TN & 62.09 & 59.87 & 73.5 & 65.99 & 11 &   1\\
Presence-TFIDF & 82.54 & 85.49 & 78.43 & 81.81 & 16 &   1\\
Presence-WFIDF & 83.54 & 84.33 & 82.42 & 83.37 & 12 &   0\\
	\hline
	\end{tabular} 
	\end{adjustbox}
	\label{table:componentClassifiers1} 
	\end{table*} 
\begin{table*}[h]
	\small
	\caption{Results of the component classifiers on the HomeBuilders using five weighting schemes}
	\begin{adjustbox}{center}
	\label{Results}
	\begin{tabular}{c c c c c c c } 
	\hline
	Matrixes & Accuracy & Precision & Recall & F-measure & Iterations & Time in seconds\\
	\hline
Frequency & 87.55 & 94.03 & 80.18 & 86.56 & 9 &   1\\
Frequency-AW & 94.5 & 95.62 & 93.27 & 94.43 & 11 &   1\\
Frequency-IDF & 94.5 & 95.79 & 93.09 & 94.42 & 8 &  1\\
Frequency-TN & 74.5 & 71.65 & 81.09 & 76.08 & 7 &   3\\
Frequency-TFIDF & 92.27 & 95.06 & 89.18 & 92.03 & 15   & 1\\
Frequency-WFIDF & 94.45 & 95.62 & 93.18 & 94.38 & 8   & 1\\
Presence & 88.59 & 95.4 & 81.09 & 87.67 & 8 &   2\\
Presence-AW & 93.95 & 95.66 & 92.09 & 93.84 & 13   & 1\\
Presence-IDF & 93.73 & 95.64 & 91.64 & 93.59 & 9  & 1\\
Presence-TN & 82.95 & 86.29 & 78.36 & 82.13 & 7 &   2\\
Presence-TFIDF & 93.86 & 95.73 & 91.82 & 93.74 & 11  & 1\\
Presence-WFIDF & 93.95 & 95.48 & 92.27 & 93.85 & 7 &  1\\
	\hline
	\end{tabular} 
	\end{adjustbox}
	\label{table:componentClassifiers2} 
	\end{table*}
%%%%%%%%%%%%%%%%%	
\paragraph{Feature reduction effect} A proper features selection usually improves the learning process in term of efficiency and effectiveness. Irrelevant features can negatively affect the learning process \citep{law2004simultaneous,zeng2009new}. Therefore, for enhancing the algorithm's performance, we conduct feature reduction via matching all adjectives and adverbs against SWN. Since we are interested in positive and negative classes only polar features are considered, and the reduction is done by removing neutral terms because they do not carry the clustering characteristic of reviews. When applying feature reduction on Airlines and HomeBuilders datasets, there are slight changes which are shown in Figure \ref{fig:FeaturesReduction}. 

\begin{figure}[]
	\centering
  \includegraphics[scale= 0.9]{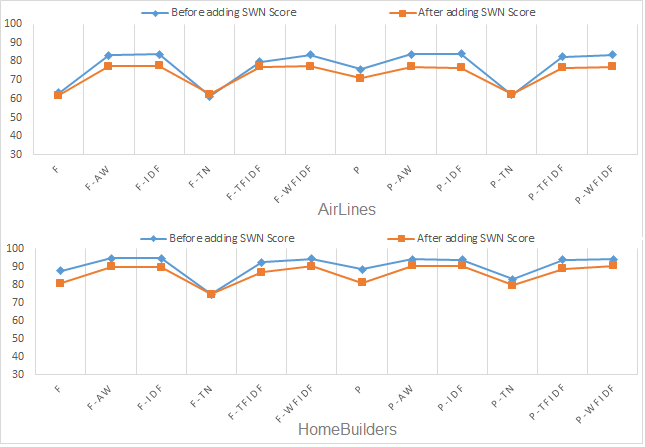}
  \caption{Adding SWN score to AriLines and HomeBuilders datasets.}
  \label{fig:SWNafterFeatureReduction}
\end{figure}

\paragraph{Sentiment scores} In Figure \ref{fig:SWNafterFeatureReduction}, the sentiment scores from SWN are added to all the matrixes. The polarity score has a negative impact on accuracy which was anticipated because the sentiment score is the average score of the synsets to which a term belongs, and the context in which a term occurs, is not considered. However, the average score is likely to correctly indicate the term polarity, that is, whether it is positive, negative or neutral. This step doubles the number of the vector space models which improves the ensemble method by promoting the groups' identification. 

\begin{figure}[]
\centering
  \includegraphics[scale= 0.9]{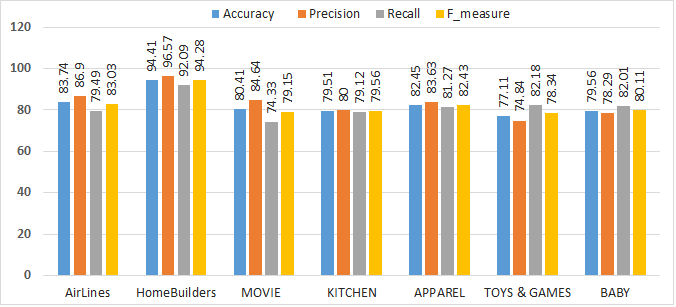}
  \caption{The performance of ACACCE on different datasets}
  \label{fig:multidomain}
\end{figure}

\paragraph{Experiments on multi-domain datasets} In this section, we present the results of ACACCE on different domains datasets. After applying the contextual analysis and constructing the matrixes, the last step is to feed the matrixes into the ensemble method, in which a document will be classified as a positive/negative instance if the majority of the component classifiers classify this document as positive/negative. The ensemble method combines 24 matrixes which are the VSMs in Tables \ref{table:componentClassifiers1} and \ref{table:componentClassifiers2} in addition to those being produced by adding SWN score. This variation will make the group interpretation more reliable because the two seeds will be classified by various data representations. In very few cases, component classifiers may incorrectly classify one of the seeds, which means both seeds will appear in one cluster. These classifiers will not be considered in the voting. An experiment on the multi-domain dataset is shown in Figure \ref{fig:multidomain}. In addition to the Airlines and HomeBuilders datasets, five sets of products' reviews (movie, kitchen, apparel, toys and games and baby) were compared. The accuracy rate is between 94.41\% and 79.56\% for six datasets except toys and games dataset where the accuracy is 77.11\%. In general, the results show that ACACCE is a domain-independent algorithm with a competitive accuracy. 			

\subsection{Discussion}
\label{Discussion} 

Table \ref {table:Evaluation} is a comparison between ACACCE and seven different classifiers. Four of them are supervised classifiers namely Support Vector Machine (SVM), Random Forest (RF), decision tree (J48), Naive Base (NB) and Multinominal Naive Base (MNB). For conducting experiments using supervised classifiers, all part of speach tags are used as a set of features and TF-IDF weight is also utilized. We also report results of a clustering based method by \citet{li2012application} on a sample of movie review dataset. In addition to that, a baseline classifier is constructed to classify a document by aggregating SWN's average scores of its adjectives and adverbs to determine the polarity. 
\begin{table*}[h]
\centering 
\caption{Evaluation}
\begin{tabular}{cccccccccc} 
\hline
Datasets	&	ACACCE	&	SVM	&	RF	&	J48	&	NB	&	MNB	&	Clustering \citep{li2012application} &	Baseline\\
	\hline
AirLines &	 83.74 & \textbf{86.4} & 85.07 & 70.93 & 73.33 & 85.33 & | & 73.13\\
HomeBuilders &	94.41 & 93.45 & \textbf{94.73} & 85.45 & 80.91 & 94.36 & | & 86.41\\
Movie &	80.41 & \textbf{83.6} & 75.6 & 68.6 & 66.4 & 75 & 77.17 - 78.33 & 66.1\\
Kitchen &	\textbf{79.51} & 77.6 & 76.4 & 69.8 & 72.2 & 77.6 & |& 70.65\\
Apparel	& \textbf{82.45} & 79.8 & 80.6 & 67 & 71.2 & 77.8 & |& 73.6\\
Toys\&Games &	77.11 & \textbf{79.6} & 79 & 70.2 & 75.8 & 76 & |& 70.0\\
Baby & \textbf{79.56} & 77.11 & 77.11 & 64.9 & 70.22 & 75.11 & |& 67.11\\
 \hline
Average & 82.45 & \textbf {82.5} & 81.21 & 70.98 & 72.86 & 80.17 & | & 72.42\\
\hline
\end{tabular} 
\label{table:Evaluation} 
\end{table*}
%%%%%%%%%%%%%%%%%%%%%%%%%%%
The results show that the average rate of the baseline classifier's accuracy is comparably low which is probably because the average score extracted from SWN does not reflect an accurate sentiment strength of a term which is mainly because the language context in which this term appears is neglected. As it can be seen in Table \ref{table:Evaluation}, The performance of ACACCE is a competitive performance compared to supervised and unsupervised classifiers. The average of the accuracy rate of ACACCE is very close to the   average rate of SVM's accuracy which is the best average rate among the compared methods. ACACCE also yields the best performances on three datasets and comparable performances on the other four datasets. The accuracy of ACAECC is enhanced by at least 2\% compared to an unsupervised method by \citet{li2012application} which is due to the contextual analysis phase, using initial polar seeds and utilize a diverse weight schemes.

The algorithm has solved the instability problem of k-means in a more efficient way. Unlike Li and Liu’s method proposed in \citep{li2012application}, this study proposes more robust and reliable method because every run on the same data the algorithm guarantees the same performance and outcome which is due to the nonstochastic centroids initialization. In addition, ACACCE provides more reliable group interpretation strategy by using assembly to classify the seeds. The advantages of the method are: (1) it is a competitive method in terms of accuracy, (2) it is stable and domain independent; and (3) it requires no human participation (i.e. unlike the supervised learning methodologies, it requires no training).

\paragraph{Research implications:}
This study has shown that SA can be effectively addressed by unsupervised clustering learning which results in a domain-independent algorithm. Our findings from experimenting on multi-domain datasets promote the idea of adopting a cluster analysis method for SA. ACACCE involves two main stages that improved the outcome; contextual analysis and an ensemble of clustering classifiers. Utilizing contextual analysis has a significant impact on the results because the language forms, which are tackled, are very common and can be strong sentiment shifters such as negation and intensifiers. 

In the ensemble learning, we have used the traditional representation of corpus where the documents are the observations and the words are the features (adjectives and adverbs). This study supports what have been suggested in previous research \citep{benamara2007sentiment} that adjectives and adverbs are the most informative parts of speech in term of sentiment analysis. However when it is binary classification only polar adjectives and adverbs are significant for learning, this is can be seen (Figure \ref{fig:FeaturesReduction}) when eliminating neutral terms which has is no significant impact on the results.
  
Ensemble method has positive implications on ACACCE. Increasing the number of diverse and accurate classifiers slightly enhances the algorithm accuracy, and this is supporting what have been stated in \citep{hansen1990neural}. More importantly, assembling is a significant strategy for group judgment as being explained in sections \ref{ClusterInterpretations} and \ref{EnsembleLearning}. The experiments' results using diverse term weighting schemes indicate that the term weighting in the entire corpus is more important compared to its weight in a particular document.

One of the research observations is that the generalization performance of ACACCE is enhanced which is a result of applying contextual analysis and using various data representation. In Table \ref{table:Evaluation}, ACACCE's performance is relatively stable when operating on different datasets compared to other algorithms. For example, ACACCE yields the higher accuracy when operating on Kitchen dataset whereas the accuracy rates of the other algorithms are comparably low when processing this dataset. This is because of the two processing phases of ACACCE where the processed text will be automatically analyzed in the contextual analysis phase, then in the second phase different data representations will be combined in the ensemble method.

This study shows that SA can be addressed by employing unsupervised clustering algorithm. k-means as a base classifier is suitable for our method because in ACACCE the cluster number is pre-defined, and k-means can process a large quantity of data in short time because it is a nonhirariacal algorithm. The k-means also being enhanced by nonrandom centroids selection which significantly increases k-means accuracy and efficiency. Selecting initial points for k-means is crucial and it was a research topic for some studies \citep{laszlo2007genetic,babu1993near,yedla2010enhancing}. 

\section{Conclusions}
\label{Conclusions}

In this article, we have discussed a completely automatic unsupervised machine learning method for sentiment analysis. The method combines an automatic contextual analysis and an ensemble of clustering classifiers. Unsupervised learning and reliability are the features that distinguish the proposed ensemble algorithm from the other work in the literature. The reliability of ACACCE is derived from the combination of the contextual analysis phase and the ensemble learning methodology. It is an unsupervised algorithm with competitive accuracy, and subsequently, it is a domain-independent classification algorithm. ACACCE solves the problem of data annotation, which is an expensive process.  

As a future work, we will consider a multi-class classification based on the sentiment strength. An enhancement can also be achieved by considering deeper contextual analysis and utilizing other weighting schemes or even other machine learning approaches.

\textbf{Acknowledgment} \\
The authors would like to acknowledge the financial support from the Iraqi Ministry of Higher Education and Scientific Research (MoHESR).
 


 \bibliographystyle{elsarticle-harv} 
\bibliography{ref}



%%\bibitem[ ()]{}

%%\end{thebibliography}
\end{document}